\def\year{2020}\relax
\documentclass[letterpaper]{article} 
\usepackage{aaai20}  
\usepackage{times}  
\usepackage{helvet} 
\usepackage{courier}  
\usepackage[hyphens]{url}  
\usepackage{graphicx} 
\usepackage{graphicx}  
\usepackage{subcaption} 
\usepackage{amsfonts} 
\usepackage{amsmath}
\usepackage[]{algorithm2e}

\newcommand{\mtx}[1]{\ensuremath{\mathbf{#1}}}
\newcommand{\vtr}[1]{\ensuremath{\mathbf{#1}}}
\newcommand{\argmin}{\mathop{\mathrm{argmin}}}    
\title{Representing Closed Transformation Paths in Encoded Network Latent Space}

\author{
  Marissa C. Connor \ and Christopher J. Rozell\\
  Department of Electrical and Computer Engineering\\
  Georgia Institute of Technology\\
  Atlanta, GA 30332 \\
  \texttt{(marissa.connor,crozell)@gatech.edu} \\
}

\begin{document}

\maketitle

\begin{abstract}
Deep generative networks have been widely used for learning mappings from a low-dimensional latent space to a high-dimensional data space. In many cases, data transformations are defined by linear paths in this latent space. However, the Euclidean structure of the latent space may be a poor match for the underlying latent structure in the data. In this work, we incorporate a generative manifold model into the latent space of an autoencoder in order to learn the low-dimensional manifold structure from the data and adapt the latent space to accommodate this structure. In particular, we focus on applications in which the data has closed transformation paths which extend from a starting point and return to nearly the same point. Through experiments on data with natural closed transformation paths, we show that this model introduces the ability to learn the latent dynamics of complex systems, generate transformation paths, and classify samples that belong on the same transformation path.
\end{abstract}

\section{Introduction}
In many applications of interest, intelligent algorithms must classify, generate, or understand natural, high-dimensional data. The manifold hypothesis states that high-dimensional data can be modeled as lying on a low-dimensional, nonlinear manifold~\cite{fefferman2016testing}. A variety of techniques have been introduced in the machine learning literature to discover various aspects of manifold structure from data, often estimating the low-dimensional nonlinear structure of high-dimensional data using neighboring points to define local geometric structure~\cite{tenenbaum2000global,roweis2000nonlinear,dollar2007learning,bengio2005non,rao1999learning,miao2007learning,culpepper2009learning}. 
\begin{figure*}[ht]

\centering
\begin{subfigure}[b]{0.38\columnwidth}
  \centering
	{\includegraphics[width=0.85\columnwidth]{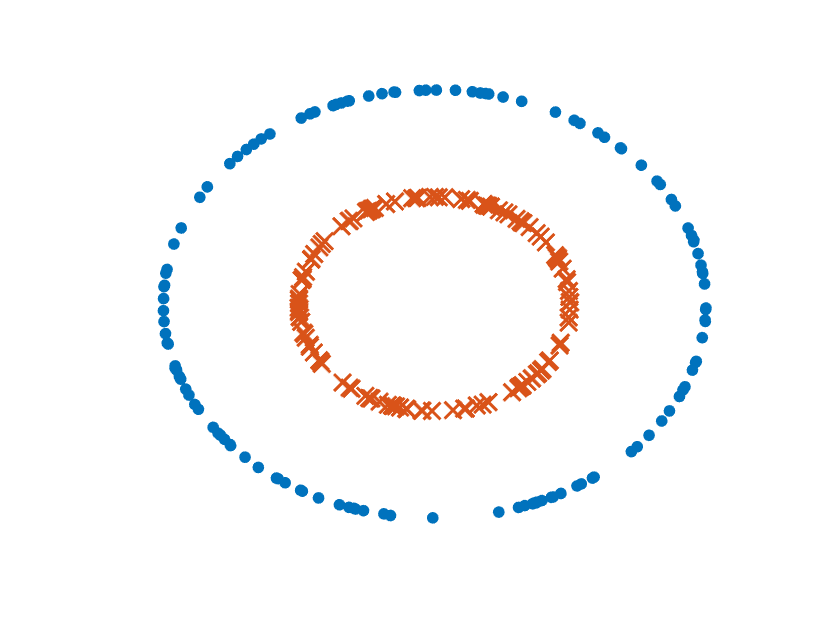}}
  \caption{}
	\label{subfig:circleTrain}
\end{subfigure}
\begin{subfigure}[b]{0.38\columnwidth}
  \centering
	{\includegraphics[width=0.85\columnwidth]{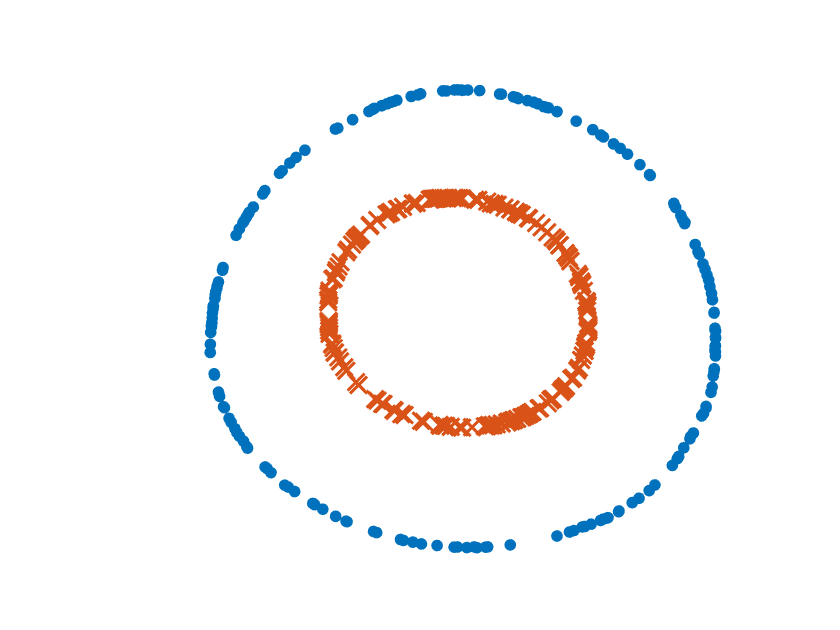}}
  \caption{}
	\label{subfig:circleLatent}
\end{subfigure}
\begin{subfigure}[b]{0.38\columnwidth}
  \centering
	\includegraphics[width=0.85\columnwidth]{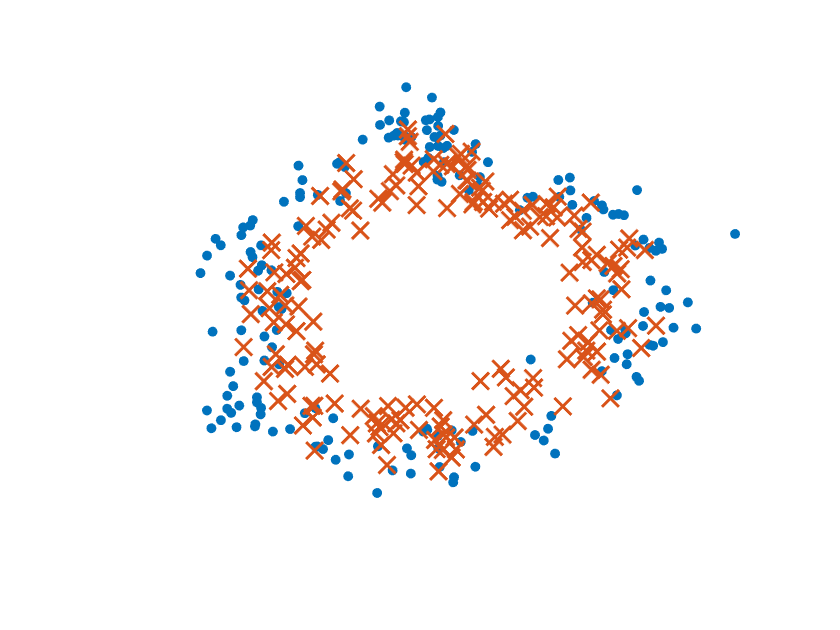}
	\caption{}
	\label{subfig:circleVAE}
\end{subfigure}
\begin{subfigure}[b]{0.38\columnwidth}
  \centering
	{\includegraphics[width=0.85\columnwidth]{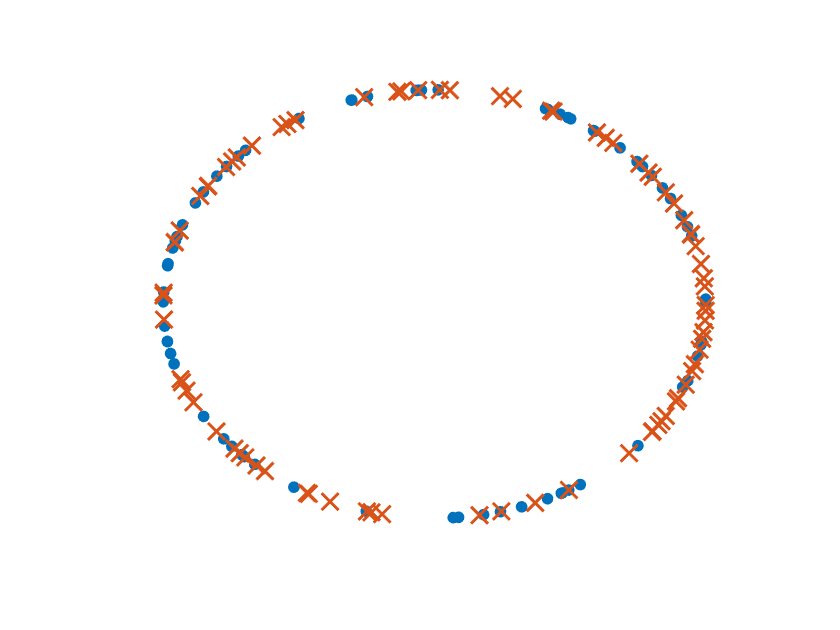}}
  \caption{}
	\label{subfig:circlehvae}
\end{subfigure}
\begin{subfigure}[b]{0.38\columnwidth}
  \centering
	{\includegraphics[width=0.85\columnwidth]{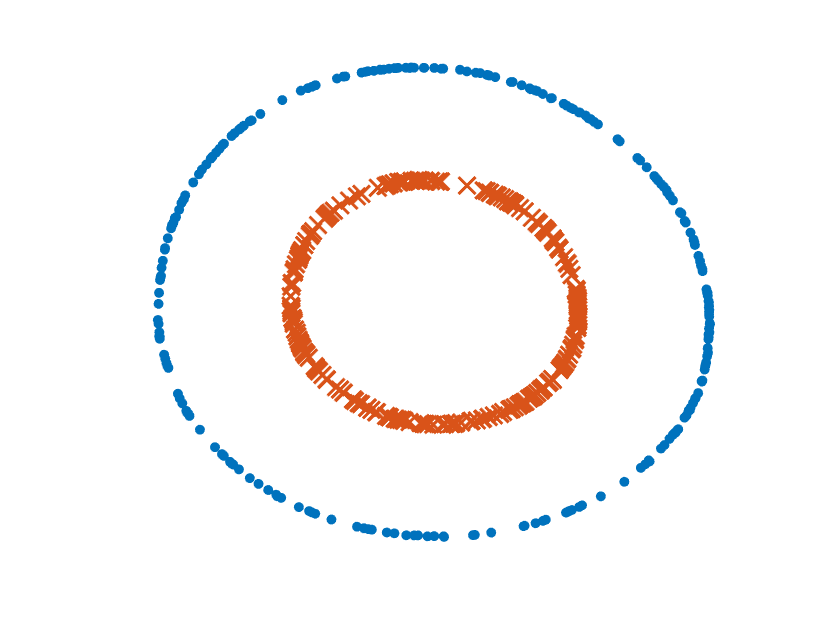}}
  \caption{}
	\label{subfig:circlemae}
\end{subfigure}

  \caption{\label{fig:vaeCompare} (a) Ground truth latent embedding. (b) Circle points embedded in the autoencoder latent space. (c) Circle points embedded in the VAE latent space. (d) Circle points embedding in the hyperspherical VAE latent space (e) Circle points embedded in the manifold autoencoder latent space }
	
\end{figure*}

Recently, deep generative models have been used to learn generator functions $f : \mathcal{Z} \rightarrow \mathcal{X}$ that map points from a low-dimensional latent space, $\mathcal{Z} \subseteq \mathbb{R}^d$,  to a high-dimensional data space, $\mathcal{X} \subseteq \mathbb{R}^D$. These generative models, which include variational autoencoders (VAEs)~\cite{kingma2013auto} and generative adversarial networks (GANs)~\cite{goodfellow2014generative}, can generate high-fidelity output samples that resemble real-world data. Generative networks encourage the latent space to fit a prior distribution, like a Gaussian, and often the latent space is described as a data manifold in which natural data transformations are defined by linear paths~\cite{radford2015unsupervised}. The assumption that transformations are represented by linear paths severely limits the complexity of transformation paths that can be continuously represented by the model. 

An example scenario where existing methods are insufficient is an application in which the data has closed transformation paths that extend from a starting point, $\vtr{z}_0$, and return to nearly the same point $\vtr{z}_N \approx \vtr{z}_0$ after $N$ steps. There are many examples of applications that fit this setting including 3D rotating objects, temporal action sequences, and natural systems governed by underlying dynamical processes. These transformations cannot be represented by linear paths in the neural network latent space but instead require a more complex representational model for latent data variations. There are two general approaches that can incorporate natural data structure into latent spaces of generative networks to represent more complex transformations. The first approach encourages geometric priors in the latent space that may better accommodate natural data variations. The second approach defines transformations using geodesic paths in the latent space rather than linear paths.

We combine both of these approaches by explicitly incorporating a learned generative manifold model into the latent space of an autoencoder. This manifold model will allow us to both learn structure associated with natural variations in the data and define nonlinear paths necessary for representing closed transformation paths. In this work, we consider data with closed transformations paths and propose an approach to learn and represent this natural manifold structure in a neural network, with the following specific contributions:
\begin{itemize}
    \item We develop a model for learning generative, nonlinear manifold operators between pairs of points in the latent space that represent natural transformations in the data. We show that this model can interpolate paths between points and extrapolate closed paths from a single starting point. 
    \item We create a network architecture that incorporates the manifold transformation operators into an autoencoder latent space. This enables us to define a latent space structure that adheres to the structure in the data itself. We show that incorporating a learned, nonlinear manifold structure in the latent space greatly improves our ability to generate full loops of closed transformation paths.
    \item We define a distance metric that determines the likelihood that two points, $\vtr{z}_0$ and $\vtr{z}_1$, lie on the same manifold given the learned manifold transformation operators. We show that this distance can be used to find new samples that exist on the same closed transformation path as a reference point.\end{itemize}

With this model, we gain the ability to learn the latent dynamics of complex systems, generate transformation paths, and classify samples that belong on the same transformation path. We apply this method to three data sets that demonstrate the importance of representing closed transformation paths: concentric circles, digit rotation, and gait sequences.

\section{Nonlinear Structure in Encoded Network Latent Space}\label{sec:background}
VAEs and GANs have gained in popularity because they are able to generate realistic data samples from the latent space distribution, $p(\vtr{z})$. Traditionally priors in the latent space are chosen for their computational simplicity rather than their compatibility with the latent structure of a given dataset, leading to the potential for an inaccurate low-dimensional representation of data variations. While efforts have been made to increase the complexity of these priors, there remain significant gaps in methods that incorporate both meaningful assumptions of geometric structure and the ability to adapt the detailed prior to the data characteristics. For example, some techniques encourage latent spaces with more complex geometric structure (e.g., hyperspheres, tori, and mixtures of Gaussians) by specifying prior distributions on the latent vectors during training, but they are non-adaptive and therefore may not match the specific data structure~\cite{tomczak2017vae,davidson2018hyperspherical,makhzani2015adversarial,falorsi2018explorations,rey2019diffusion}. On the other hand, some techniques train a recurrent neural network on latent vectors to model sequences of dynamical systems~\cite{sussillo2016lfads}, but do not incorporate prior knowledge of geometric structure likely present in the data.

In lieu of implementing a structured geometric prior, some techniques incorporate natural data structure into a latent space by computing geodesic paths between points using an estimated Riemannian metric~\cite{arvanitidis2017latent,chen2017metrics,shao2018riemannian}. Similar to our main aims, these techniques identify situations in which linear paths are not effective for representing transformations and aim to estimate more accurate nonlinear manifold paths between points. However, these techniques estimate individual paths by inferring the shortest distance between two points given the computed Riemannian metrics. This does not specify a general structure for the latent space or enable efficient extrapolation of transformation paths. 

We develop a \textit{manifold autoencoder} which incorporates a generative manifold model into the latent space of an autoencoder. Unlike the methods above which encourage the latent space to model a specified prior, our method learns the latent structure from the data. Additionally, our manifold model is made up of learned generative operators that can easily interpolate paths between and extrapolate paths from arbitrary points in this space. 

\subsection{Closed Transformation Paths}

Fig.~\ref{fig:vaeCompare} provides an example of why it is necessary to address closed transformation paths directly and why we are using an autoencoder model to define our encoded latent space. The training data for these plots are 20-dimensional features that are mapped from the two-dimensional, ground truth latent space in Fig.~\ref{subfig:circleTrain}. The points are embedded onto two circular manifolds. Fig.~\ref{subfig:circleLatent} shows the embedding from a learned autoencoder with a two-dimensional latent space. This autoencoder effectively represents the concentric circular latent structure of the data. By contrast, the embedding from the trained VAE mixes the points from the two manifolds because the Gaussian prior encourages all the points to be centered at the origin (Fig.~\ref{subfig:circleVAE}). This Gaussian prior is clearly not appropriate for data with closed transformations paths. To address this manifold mismatch, the hyperspherical VAE implements a hyperspherical prior into the latent space~\cite{davidson2018hyperspherical}. The hyperspherical structure in the latent space introduces the possibility of creating closed paths. However, this technique does not effectively define individual object manifolds but rather combines the manifolds of all training objects onto the same hyperspherical space. Fig.~\ref{subfig:circlehvae} shows this effect because points from the two data manifolds are embedded on the same circle in the two-dimensional latent space. 

To address the issues with manifold mismatch and closed transformation paths, our manifold autoencoder learns the structure from the data itself and adapts the latent space of an autoencoder to fit that learned structure. Fig.~\ref{subfig:circlemae} shows the embedding of the points from our manifold autoencoder which will be described in detail in Section~\ref{subsec:manigeo}. The points are adapted to fit the circular structure of the data while maintaining two concentric circles.

\section{Methods}

\subsection{Learning Natural Manifold Geometry}\label{sec:lieOpt}
To characterize transformations in a latent space, we need to define a transformation operator, $\mtx{T}$, that can generate the transformations between two data points $\vtr{z}_1 = \mtx{T}\vtr{z}_0$. This transformation belongs to a family of operators $\mathcal{T}$ (i.e., $\mtx{T} \in \mathcal{T}$) which is endowed with group structure. Specifically, i) there exists a group operation $\circ$ to combine transformations; ii) if $\mtx{T}_0, \mtx{T}_1 \in \mathcal{T}$ then $\mtx{T}_0 \circ \mtx{T}_1 \in \mathcal{T}$; iii) the operators are associative; iv) there exists an identity operator; and v) there exists an inverse element for each operator $\mtx{T} \in \mathcal{T}$. This continuous transformation group is termed a Lie group~\cite{boothby1986introduction}. 

A Lie group of transformation operators in the latent space defines transformations that describe the natural variations in $\mathcal{Z}$. Thus these operators can be used to compactly represent the latent manifold surface, $\mathcal{M} \subset \mathcal{Z}$ around a latent vector, $\vtr{z}_0$: $\mathcal{M} = \{\vtr{z} \in \mathcal{Z} | \vtr{z} = \mtx{T}\vtr{z}_0, \mtx{T} \in \mathcal{T}\}$.

Each of these transformation operators represents a continuous transformation and can be formulated with the infinitesimal transformation: $\mtx{T} = (\mtx{I} + \mtx{\Psi}\Delta c)$ where $\mtx{T}$ is a small offset from the identity matrix. The real number $c \in \mathbb{R}$ parameterizes $\mtx{T}$ with a matrix $\mtx{\Psi}$ which is an operator for the transformation group. Transformed latent vectors are defined with the infinitesimal transformation as $\vtr{z}_{\Delta c} = \mtx{T}\vtr{z}_0 = (\mtx{I} + \mtx{\Psi}\Delta c)\vtr{z}_0$. As $\Delta c \rightarrow 0$, the previous equation becomes $\frac{\delta \vtr{z}}{\delta c} = \mtx{\Psi}\vtr{z}$. This dynamical system has the well-known solution $\vtr{z}_c = \mathrm{expm}(\mtx{\Psi} c)\vtr{z}_0$~\cite{rao1999learning,miao2007learning}.

We assume that the manifold in the latent space is defined by a finite set of transformation operators $\{\mtx{T}_1, \mtx{T}_2, ... \mtx{T}_M\}$, each parameterized by their own coefficient, $c_m$, which specifies the contribution of the individual operator to the given transformation. Therefore a transformation $\mtx{T}$ can be represented by a weighted combination of transformation dictionary elements, $\mtx{\Psi}_m$

\begin{equation}
    \mtx{T} = \mathrm{expm}\left(\sum_{m=1}^M{\mtx{\Psi}_mc_m}\right).
\end{equation}

This defines a generative transformation model that characterizes the manifold $\mathcal{M}$ by a set of transformations operators $\mtx{\Psi}_m$. Following previous conventions~\cite{culpepper2009learning}, we call the transformation dictionary elements $\mtx{\Psi}_m$ \textit{transport operators}. The relationship between points in the latent space is defined as:

\begin{equation}
\vtr{z}_1 = \mathrm{expm}\left(\sum_{m=1}^M{\mtx{\Psi}_mc_m}\right)\vtr{z}_0 + \vtr{n}, \label{eqn:dynamics}
\end{equation}
where $\vtr{n}$ represents additive noise.

Using the relationship between points in (\ref{eqn:dynamics}), a probabilistic generative model can be written that allows efficient inference. This model assumes a Gaussian noise model, a Gaussian prior on the transport operators $\mtx{\Psi}$ (model selection), and a sparsity inducing prior on the coefficients $\vtr{c}$ (model regularization).  The resulting negative log posterior for the model is given by
\begin{equation} \label{eq:objFun}
\begin{split}
E_{\Psi} = \frac{1}{2}\left\|\vtr{z}_1 - \mathrm{expm}\left(\sum_{m=1}^M{\mtx{\Psi}_mc_m} \right) \, \vtr{z}_0\right\|_2^2 \\
+ \frac{\gamma}{2}\sum_m\|\mtx{\Psi}_m\|_F^2 +\zeta\|\vtr{c}\|_1,
\end{split}
\end{equation}
where $\|\cdot\|_F$ is the Frobenius norm. The transport operator dictionary elements are learned using pairs of neighboring points in the latent space. The selection process for these point pairs depends on the application. If the application involves temporal data, neighboring points are defined as points that are a few time steps apart. If the data does not include temporal sequences, there are several other methods for selecting point pairs during training including finding points that are close together in an embedding space, points that share similar attribute labels, points that are identified as similar through human input, and points that are close in a neural network feature space.

\begin{figure*}[t]
\centering

\includegraphics[width=1.98\columnwidth]{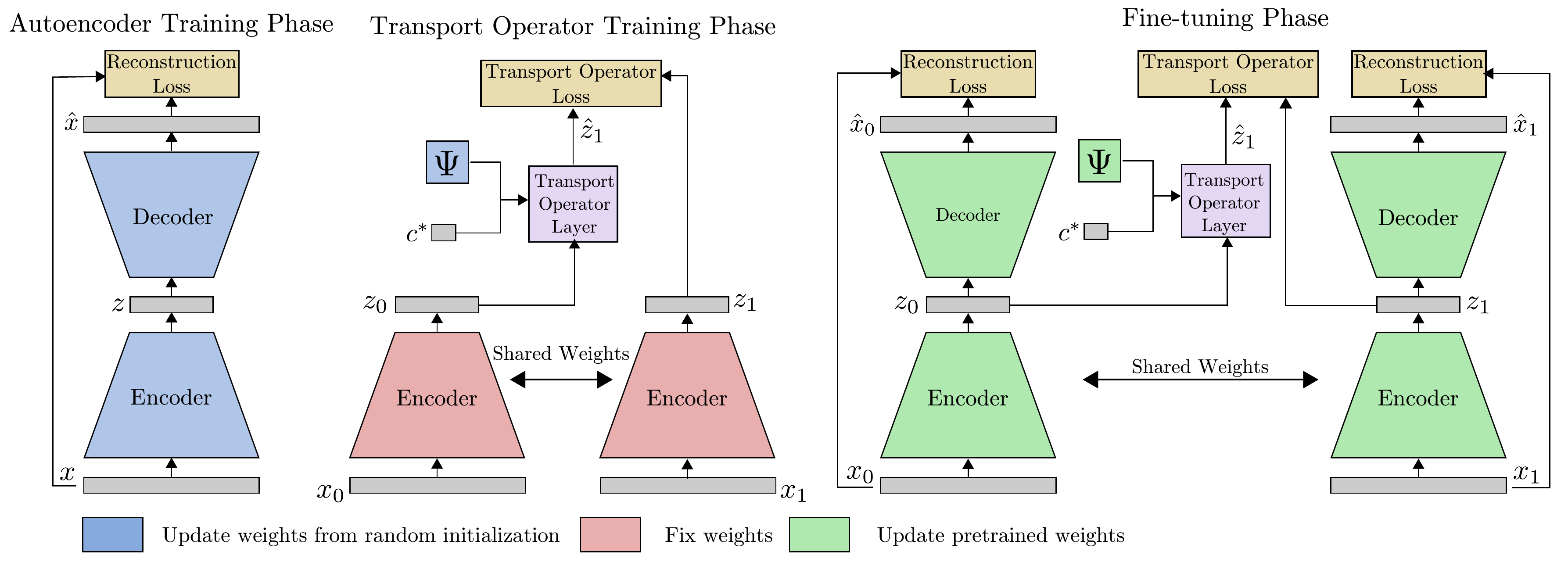}

	\caption{\label{fig:TrainDiagram} Visualization of the three phases of training the manifold autoencoder.}

\end{figure*}

Using an established unsupervised algorithm~\cite{culpepper2009learning}, the transport operator dictionary can be learned by using pairs of neighboring points to perform alternating minimization with the objective in~(\ref{eq:objFun}). Specifically, for a given batch of point pairs, we alternate between inferring the coefficients using conjugate gradient descent given a fixed dictionary of transport operators and taking a step in the gradient direction of the dictionary elements. Note that the Frobenius norm term on the dictionary elements aids in identifying the transport operators that are needed to represent natural transformations. If a dictionary element is not used to estimate transformations between point pairs in a training batch, this term reduces the magnitude of that dictionary. 

Once the operators are learned, a transformation can be defined entirely by the set of coefficients, $\vtr{c}$, used to control the learned operators, $\mtx{\Psi}$. To specify a transformation between two points, the coefficients can be inferred between those points: $\vtr{c}^* =  \argmin_\vtr{c}E_{\Psi}$ and used to characterize an operator $\mtx{A} = \sum_{m=1}^M{\mtx{\Psi}_mc^*_{m}}$ that can be applied to a starting point, $\vtr{z}_0$: $\vtr{z}_t = \mathrm{expm}(\mtx{A}t)\vtr{z}_0$. This data generation can be used to interpolate between those points, extrapolate beyond those points, or transfer transformations to new points.

\subsection{Incorporating Manifold Geometry into Latent Space}\label{subsec:manigeo}
We develop a three phase approach for training the manifold autoencoder which incorporates transport operators that represent closed transformation paths into the latent space of an autoencoder. See Fig.~\ref{fig:TrainDiagram} for a visualization of the phases.
\begin{itemize}
    \item \textbf{Autoencoder Training Phase}: Train the autoencoder on input data.
    \item \textbf{Transport Operator Training Phase}: Fix the autoencoder weights and train the transport operators on pairs of samples from the encoded latent space as described in Section~\ref{sec:lieOpt}
    \item \textbf{Fine-tuning Phase}: Fine-tune the autoencoder network weights and transport operators simultaneously. 
\end{itemize}

\begin{figure*}[ht]

\centering
\begin{subfigure}[b]{0.8\columnwidth}
  \includegraphics[width=0.99\columnwidth]{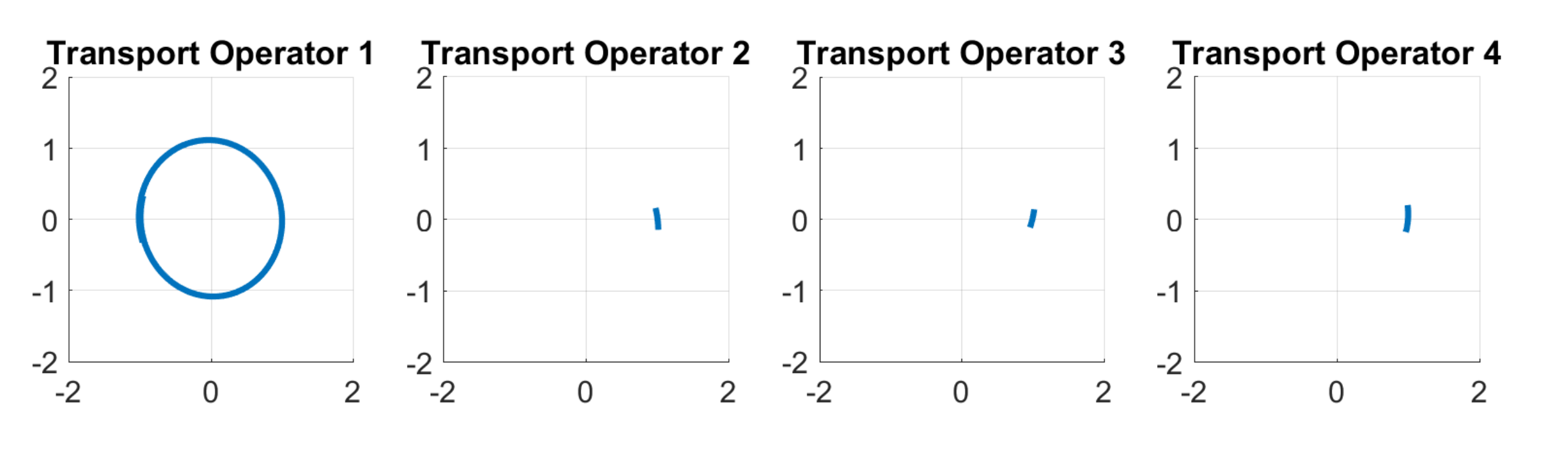}

	\caption{}\label{subfig:circleTransOpt}
\end{subfigure}
\begin{subfigure}[b]{0.58\columnwidth}
  \centering
	\includegraphics[width=0.94\columnwidth]{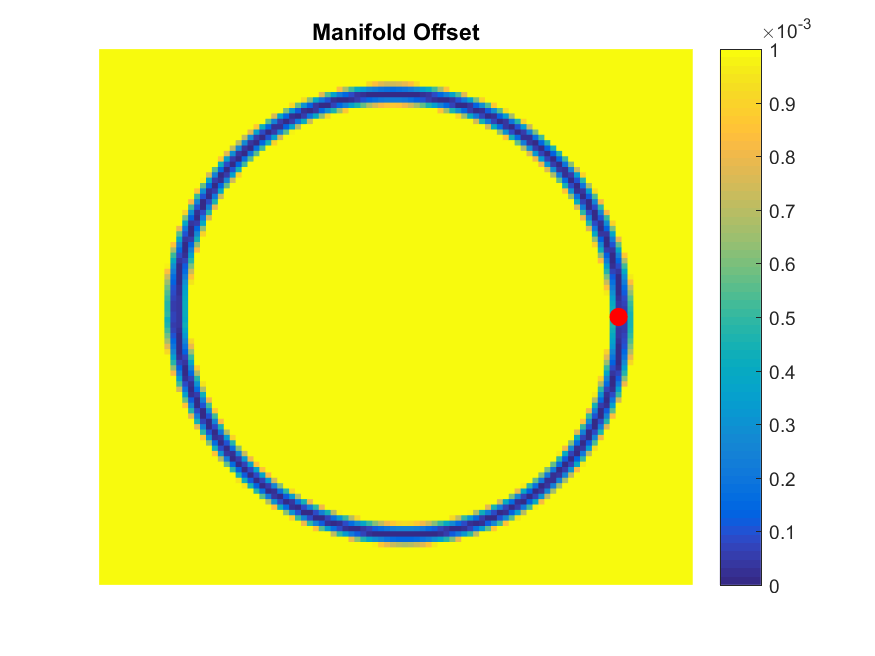}
  \caption{}
	\label{subfig:offsetHeat}
\end{subfigure}
\begin{subfigure}[b]{0.58\columnwidth}
  \centering
	\includegraphics[width=0.94\columnwidth]{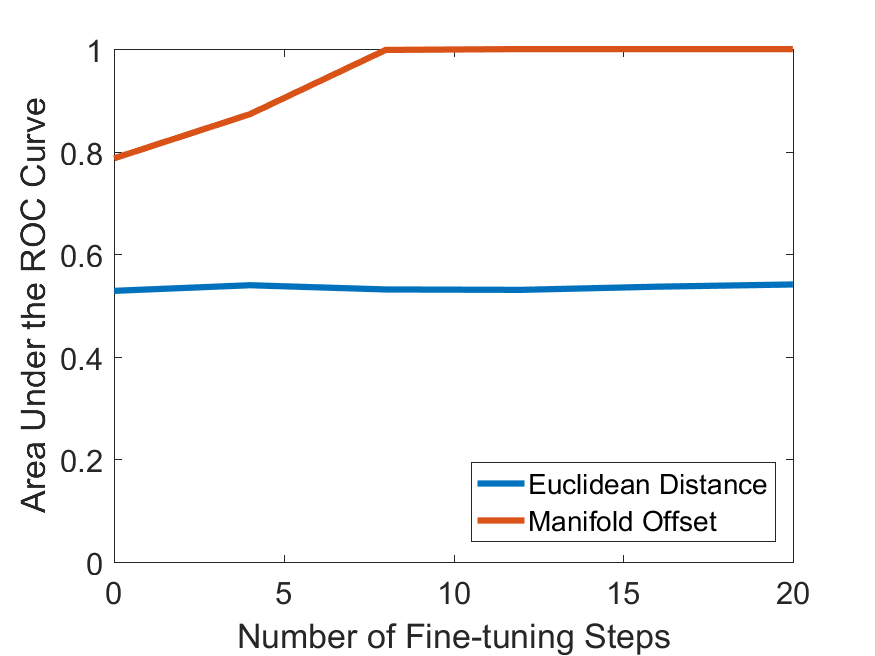}
	\caption{}
	\label{subfig:circleAUC}
\end{subfigure}
	
  \caption{\label{fig:circleDistTest} (a) Orbits of transport operators learned in the neural network latent space. (b) Heat map of the manifold offset distance from the red reference point to every point on the grid. The manifold offset distance is small on the 1D circular manifold on which the reference point resides. (c) AUC curve for classification of points on the same or different circles during fine-tuning process. As fine-tuning progresses, the manifold offset distance becomes better able to separate samples from the inner and outer circles. }
	
\end{figure*}

To incorporate the transport operators into the network architecture, we create a new transport operator layer that applies a transformation defined by the current dictionary of operators, $\mtx{\Psi}$, and a set of coefficients, $\vtr{c}$, to an input latent vector, $\vtr{z}_{in}$: $\vtr{z}_{out} = \mathrm{expm}\left(\sum_{m=1}^M{\mtx{\Psi}_mc_m}\right)\vtr{z}_{in}$. 

We begin with the autoencoder training phase in which we train an autoencoder with a reconstruction loss. We then progress to the transport operator training phase in which we fix the network weights and train the transport operators as described in Section~\ref{sec:lieOpt}. Prior to the fine-tuning phase, we identify relevant transport operators and eliminate those not being used by defining a magnitude threshold and eliminating the dictionary elements whose magnitudes are below the threshold. Note that this built-in selection of dictionaries allows us to initially overestimate the number of dictionaries, $M$, and rely on the model to identify the number necessary for representing the transformations.

During the fine-tuning phase, we simultaneously update the transport operators and the autoencoder weights. The dictionary elements are updated with the same objective function as in Eq~\ref{eq:objFun}. The network weight, $\mtx{\phi}$, updates are supervised by both reconstruction losses and a transport operator loss:
\begin{equation} \label{eq:objPhi}
E_{\phi} = \left\|\vtr{x}_0 - \hat{\vtr{x}}_0\right\|_2^2 + \left\|\vtr{x}_1 - \hat{\vtr{x}}_1\right\|_2^2 + \lambda E_{\Psi}.
\end{equation}
As with transport operator training, fine-tuning the network weights and dictionary elements involves alternating between inferring transformation coefficients between samples and taking gradient steps on the dictionaries and network weights. Algorithm 1 shows the pseudo-code for the training procedure during the fine-tuning phase.

\begin{algorithm}
 \KwData{Training samples $\vtr{x} \in \mathcal{X}$, pretrained dictionaries $\mtx{\Psi}$, pretrained network weights $\mtx{\phi}$}
 \KwResult{Fine-tuned operator dictionary elements $\mtx{\Psi} $ and network weights $\mtx{\phi}$ }
 \For{$i = 0,....,N$}{
  Select a batch of point pairs $\vtr{x}_0$ and $\vtr{x}_1$\;
  Encode point pairs to get $\vtr{z}_0$ and $\vtr{z}_1$\;
  \For{$j = 0,....,\mathrm{num\_pairs}$}{
    Initialize $\vtr{c}$  as $c_m \sim \mathrm{Unif}[0,1]$\;
    Fix $\vtr{c}$ to $\vtr{c}^* =  \argmin_\vtr{c}E_{\Psi}$\;
    $\mtx{\Psi} = \mtx{\Psi} - \eta\frac{\delta E_{\Psi}}{\delta \Psi}$\;
    $\mtx{\phi} = \mtx{\phi} - \zeta\frac{\delta E_{\phi}}{\delta \phi}$\;
  }
 }
 \caption{Fine-tuning of network weights and transport operators}
\end{algorithm}

To visualize the transport operators learned in the latent space, we revisit the dataset used in Fig.~\ref{fig:vaeCompare} and train an autoencoder with a two-dimensional latent space on 2D circular points that are mapped into a 20-dimensional space. Fig.~\ref{subfig:circleLatent} shows these points encoded in the latent space after the autoencoder training phase. Point pairs for transport operator training are created by randomly selecting $\vtr{z}_0$ from the training set and selecting a $\vtr{z}_1$ that is one of the 20 nearest neighbors of $\vtr{z}_0$. Fig.~\ref{subfig:circleTransOpt} shows the orbits of the four transport operators after the transport operator training phase. These plots are generated by applying a single learned operator as it evolves over time to a starting point in the latent space, $\vtr{z}_0$:  $\vtr{z}_t = \mathrm{exp}(\mtx{\Psi}_mt)\vtr{z}_0$, $t = 0,....,T$. The number of transport operators started at $M=4$ but the Frobenius norm term in (\ref{eq:objFun}) reduced the magnitude of the unused dictionary elements to nearly 0, resulting in only the rotational operator (transport operator 1). Finally, the fine-tuning phase adjusts both the transport operators and the latent space to accommodate one another. Fig.~\ref{subfig:circlemae} shows the embedded points after fine-tuning. These points have a more clearly circular structure than the initial embedding points in Fig.~\ref{subfig:circleLatent}.

\subsection{Manifold Offset Distance}\label{sec:distMet}
The learned transport operators define motion on class manifolds and constrain possible directions of transformations. We can use the operators to define a distance that indicates the likelihood that two latent vectors lie on the same manifold path. We call this distance the \textit{manifold offset distance} and it indicates the how well a point $\vtr{z}_1$ can be estimated from a starting point $\vtr{z}_0$ using the learned operators $\mtx{\Psi}$. The manifold offset distance is defined as: 
\begin{equation}
d_{\mbox{offset}} = \left\|\vtr{z}_1 - \mathrm{expm}\left(\sum_{m=1}^M{\mtx{\Psi}_mc^*_{m}}\right)\vtr{z}_0\right\|_2^2,
\end{equation}
 with the inferred coefficients $\vtr{c}^* =  \argmin_\vtr{c}E_{\Psi}$. 

We use the concentric circle data set and the transport operators shown in Fig.~\ref{subfig:circleTransOpt} to intuitively understand the usefulness of the manifold offset distance. Fig.~\ref{subfig:offsetHeat} shows a heat map of the manifold offset distances from the red reference point. The manifold offset is very small for the points that exist on the same 1D circular manifold as the reference point and much larger for points off of this circular manifold.

We also use the manifold offset distance to perform a simple classification task to showcase the usefulness of the fine-tuning phase of training. The classification task is to determine whether two points are on the same circle in the concentric circle data set or different circles (see Fig.~\ref{subfig:circleTrain}). To perform this classification, we compute the manifold offset distances between two points and use a binary classifier on the distances to determine whether two points are from the same circle or different circles. We quantify the performance using the area under a receiver operator characteristic (ROC) curve. The area under the curve (AUC) is a measure of class separability that spans from 0 to 1 with 1 indicating perfect separability. Fig.~\ref{subfig:circleAUC} shows the evolution in the AUC during the fine-tuning phase. We compare the AUC using the manifold offset distance and the Euclidean distance. This plot shows that the classification is no better than chance when using the Euclidean distance between points. By contrast, there is high AUC using the manifold offset distance and an increase in AUC as fine-tuning progresses. This provides evidence that the fine-tuning is structuring the latent space so the transport operators more accurately fit the encoded data. 

\section{Experiments}

In this section, we analyze the performance of our transport operator model on two datasets that contain closed transformation paths: rotated MNIST digits and gait sequences from the CMU Graphics Lab Motion Capture Database.\footnote{CMU Graphics Lab Motion Capture Database found here: http://mocap.cs.cmu.edu/} We demonstrate the usefulness of the manifold autoencoder for transformation extrapolation and manifold identification. 

\begin{figure*}[ht]

\centering
\begin{subfigure}[b]{0.96\columnwidth}
  \centering
	{\includegraphics[width=0.99\columnwidth]{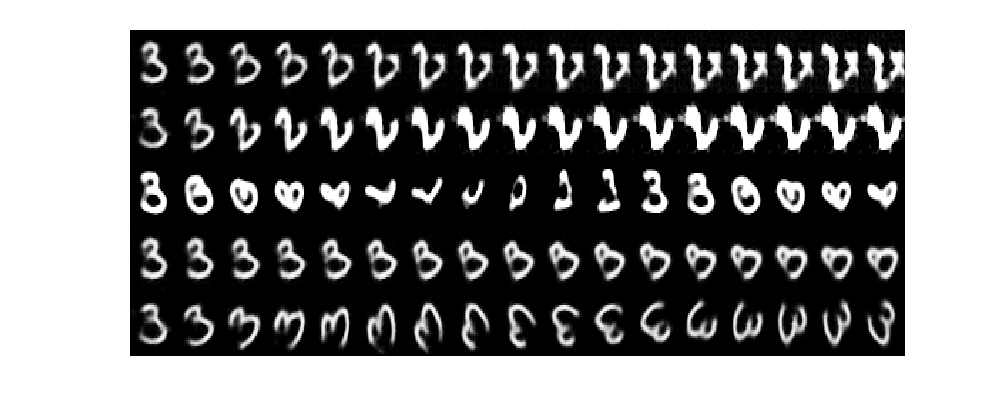}}
  \caption{}
	\label{subfig:rotTransOpt3}
\end{subfigure}
\begin{subfigure}[b]{0.96\columnwidth}
  \centering
	{\includegraphics[width=0.99\columnwidth]{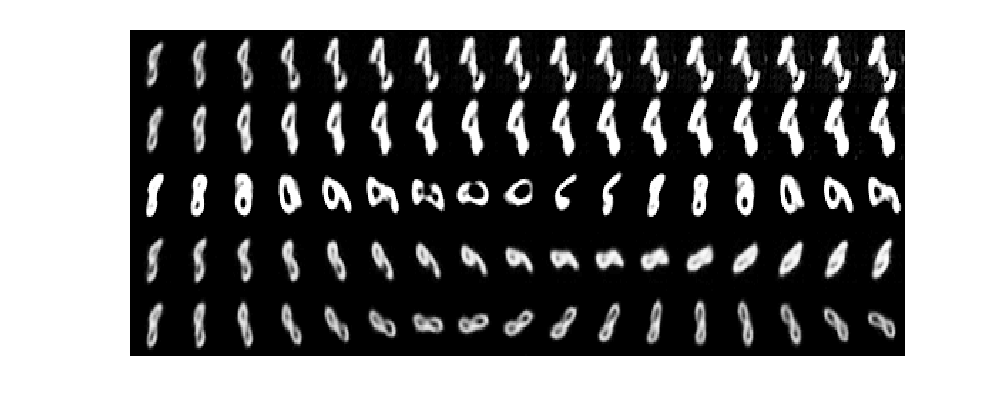}}
  \caption{}
	\label{subfig:rotTransOpt8}
\end{subfigure}
	
  \caption{\label{fig:rotTransOpt} Two examples of extrapolated rotations in the autoencoder latent space. Row 1: Extrapolated linear path in latent space prior to fine-tuning. Row 2: Extrapolated linear path in latent space after fine-tuning. Row 3: Extrapolated hyperspherical VAE path Row 4: Extrapolated transport operator path prior to fine-tuning. Row 5: Extrapolated transport operator path after fine-tuning. }
	
\end{figure*}

\subsection{Rotated MNIST Digits}
We train our autoencoder and transport operators using a subset of 50,000 images from the MNIST training set~\cite{lecun1998gradient}. In the autoencoder training phase, we train on batches of rotated MNIST digits. During the transport operator training phase, we generate pairs of rotated training digits. The first image in the pair is a digit image rotated to an angle, $\theta$, between 0 and $350^{\circ}$ degrees and the second image is that same digit image rotated to $\theta + 6^{\circ}$. The training results in one operator with a much higher magnitude than the others. Therefore, during the fine-tuning phase, we only use the single high-magnitude operator in the transport operator layer. We compare our technique against a hyperspherical VAE trained on rotated digit data. This network was trained using the network architecture and the loss functions in the code provided by the authors~\cite{davidson2018hyperspherical}.  In order to define paths in the hyperspherical VAE latent space, we compute geodesic paths on a hypersphere~\cite{Ber17}. See the appendix sections~\ref{subsec:trainParam} and~\ref{sec:hvae} for more details of the network architecture and training process.

Fig.~\ref{fig:rotTransOpt} shows a visualization of extrapolated paths in the latent space. These paths are defined by estimating the transformations between latent representations of pairs of digits with $1^{\circ}$ of rotation between them and extrapolating the transformation to estimate the full rotation path. To generate the visualizations, we i) encode a starting image into the latent space: $\vtr{z}_0 = g(\vtr{x}_0)$; ii) apply the transformation to the the starting point; and iii) decode the image output: $\hat{\vtr{x}}_t = f(\vtr{z}_t)$. This figure shows that the linear paths produce a small amount of rotation but quickly transform the appearance of the digit itself. The path in the hyperspherical VAE latent space also induces a small rotation, but as the path extends, the appearance of the original digit transforms. This is another example of how the single hyperspherical latent space is not effective for representing large, identity-preserving transformations.

The transport operator before fine-tuning extrapolates rotation to about $45^{\circ}$ before the digit appearance significantly changes. The transport operators resulting from the fine-tuning phase can generate full, 360-degree rotations with only small changes to the digit appearance. This makes it clear that the fine-tuning phase is needed to adjust the network weights and transport operators to represent the full closed transformation paths.

\begin{figure}[t]
\centering

\includegraphics[width=0.8\columnwidth]{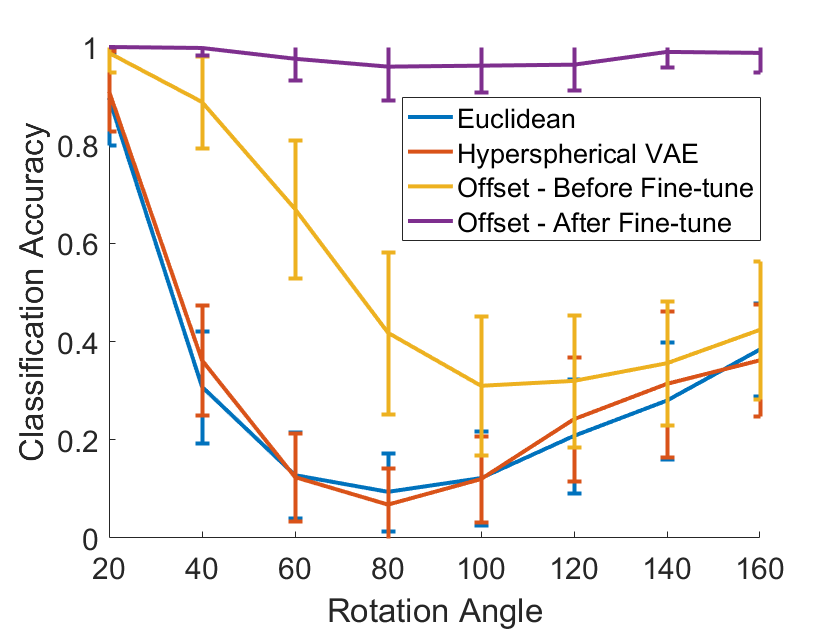}

	\caption{\label{fig:kNNFig} Nearest neighbor classification accuracy of rotated test digits. }

\end{figure}

The manifold offset distance can be used to identify samples on the same manifold transformation path. The rotated paths of individual images represent 1D manifolds in this setting and our task is to identify points on the same rotated digit manifolds using the manifold offset distance. This is shown through nearest-neighbor classification in the latent space. We define 10 training samples, the latent representation of one image for each digit class at zero degrees of rotation. The testing samples are latent representations of each of the training samples rotated to different angles. We classify each rotated sample using the class label of the nearest neighbor in the training set. Fig.~\ref{fig:kNNFig} shows the nearest neighbor classification accuracy at different rotation angles using the Euclidean distance, the geodesic distance in the hyperspherical VAE latent space, and manifold offset distance before and after fine-tuning for 50 trials. This shows that the manifold offset distance prior to fine-tuning, the Euclidean distance, and the geodesic distance in the hyperspherical VAE are poor measures for nearest neighbor classification at large rotation angles. However, after fine-tuning, the manifold offset distance is very effective for identifying points on the same rotational manifold.

\begin{figure*}[t]

\centering
\begin{subfigure}[b]{1.5\columnwidth}
  \centering
	{\includegraphics[width=0.99\columnwidth]{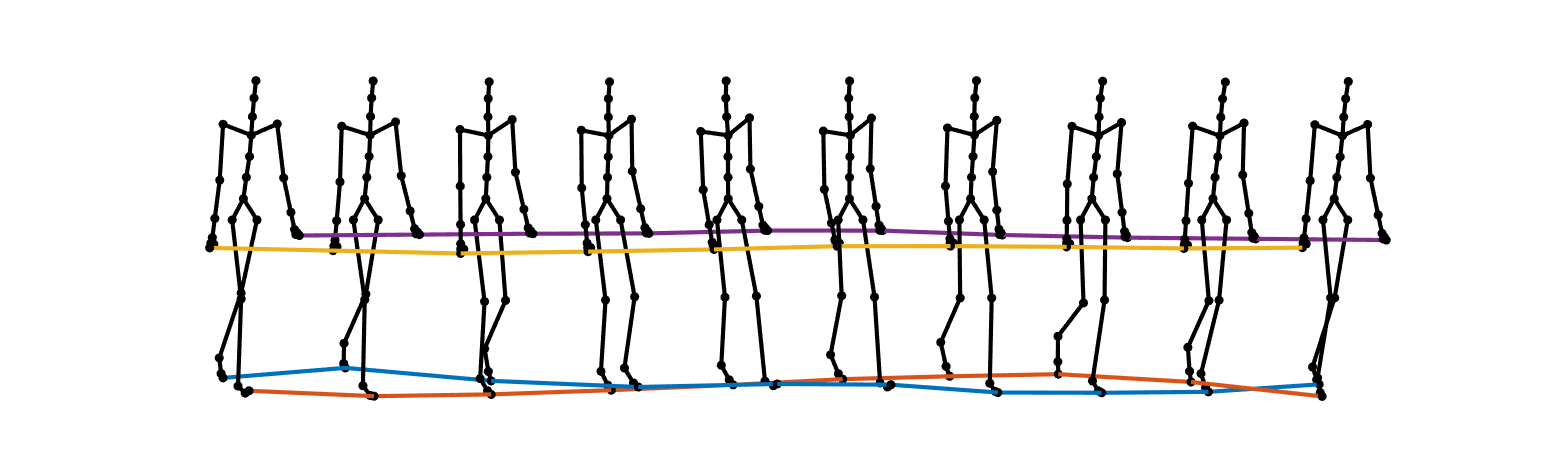}}
  \caption{}
	\label{fig:walkTransOpt}
\end{subfigure}

\begin{subfigure}[b]{0.76\columnwidth}
  \centering
	{\includegraphics[width=0.90\columnwidth]{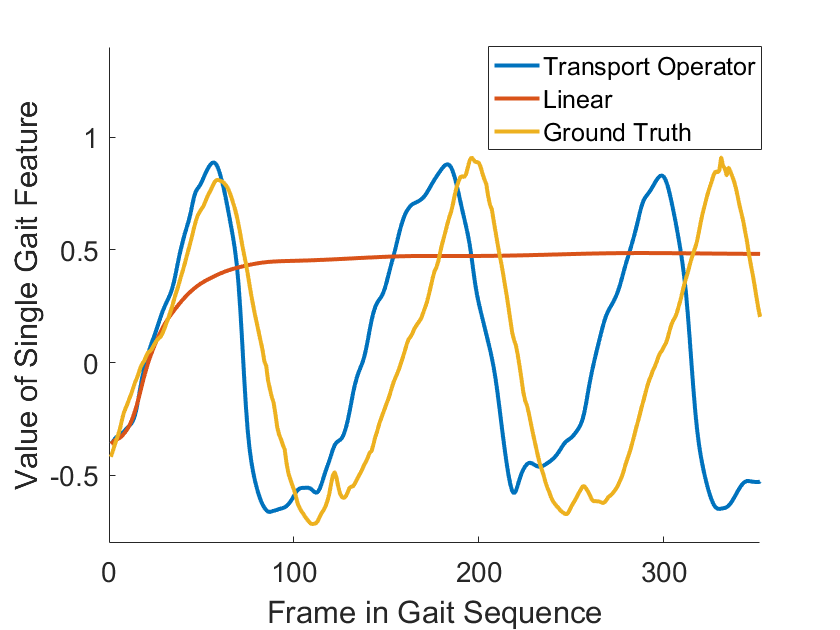}}
  \caption{}
	\label{fig:gaitFeat}
\end{subfigure}
\begin{subfigure}[b]{0.76\columnwidth}
  \centering
	{\includegraphics[width=0.90\columnwidth]{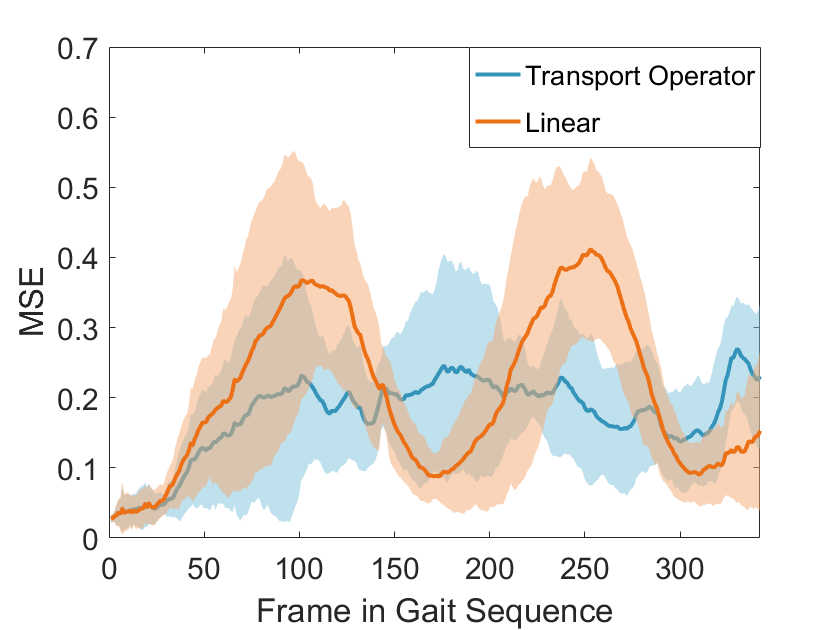}}
  \caption{}
	\label{fig:gaitEst}
\end{subfigure}

  \caption{\label{fig:walkData} (a) The effect of applying a learned operator which generates a full gait sequence. (b) Example of one feature from the data space as it progresses during an extrapolated gait sequence. (c) The error in between the estimated gait sequences and the ground truth gait sequences. The estimated sequences are extrapolated from a starting point in the latent space.}
	
\end{figure*}

These experiments highlight key benefits of our method which encourages nonlinear manifold structure in the autoencoder latent space. First, we show how the learned transport operators can be used to extrapolate full closed transformation paths from arbitrary starting points. Second, we demonstrate the advantage of incorporating the manifold model into the training of the network itself to fine-tune the encoder mapping. Finally, we show the manifold offset distance can be used to identify pairs of points on the same transformation manifold.

\subsection{Gait Sequences}

The CMU Graphics Lab Motion Capture Database includes human walking sequences that were recorded by a motion capture system. The motion data is represented through 62-dimensional feature vectors that specify locations and orientations of joints. We use the preprocessing procedure discussed in~\cite{chen2015efficient} which converts the 62-dimensional features into 50-dimensional features. We train on walking sequences 1-16 from subject 35. Walking data from this subject is abundant and this subject has been widely used for gait analysis~\cite{chen2017metrics,chen2015efficient,taylor2007modeling}.

We train an autoencoder with a five-dimensional latent space using features from the gait sequences. The point pairs used to train the transport operators are composed of features from two frames that are separated by a five frame interval. We begin with 10 transport operator dictionary elements but we eliminate all but three operators due to their low magnitudes after the transport operator training phase. Each of these three operators induces a continuous gait sequence with some difference between the mechanics of the movement. See the appendix sections~\ref{subsec:trainParam} and~\ref{sec:joinGait} more details on the training process and an analysis of how the three highest magnitude operators are jointly used.

Fig.~\ref{fig:walkTransOpt} shows an example gait sequence generated using the learned operator with the highest magnitude. This operator can represent the full loop of a gait transformation. To quantify our ability to estimate ground truth gait paths, we begin at a latent vector that encodes a neutral standing pose and use both a learned manifold transport operator and a linear path to estimate the ground truth gait sequence. We estimate the transformation between latent representations of features that are six frames apart in the gait sequence. We then use the estimated transformations to extrapolate full gait sequence paths in the latent space. We test gait estimates on test sequences 30 -34. Fig.~\ref{fig:gaitFeat} shows the extrapolated paths of a single feature in a gait sequence that is decoded from the extrapolated latent paths. We see that the linear path can effectively estimate the ground truth gait sequence in the initial frames but it eventually plateaus at a meaningless value. In comparison, the transport operator extrapolated path can match the ground truth for several gait sequences.  Fig.~\ref{fig:gaitEst} shows the mean squared error between the ground truth gait sequence and the estimated gait sequences for five test walking sequences. The error in the transport operator paths is largely due a mismatch in the speed of the extrapolated and the ground truth gait sequences.

\section{Conclusion}

We have shown that incorporating a manifold model in the latent space of an autoencoder network enables us to learn generative representations of natural data variations that define closed transformation paths. We developed a model for learning manifold operators in the latent space, incorporated these operators into the network architecture of an autoencoder, and defined a manifold offset distance. We showed the power of this approach for extrapolating paths and classifying samples using a rotated MNIST dataset and gait sequences. Future work will extend this concept of learning natural manifold transformations in a neural network latent space to datasets with unlabeled natural transformations that may not represent closed paths. This extension will employ the same methods presented here but with new techniques for selecting training point pairs on the same manifold and new tests for determining how effectively the learned operators match the natural, unlabeled transformations. 

\section{ Acknowledgments}
This work is partially supported by NSF CAREER award CCF-1350954, ONR grant number N00014-15-1-2619 and AFRL contract number FA8750-19-C-0200. The data used in this project was obtained from mocap.cs.cmu.edu. The database was created with funding from NSF EIA-0196217.

\section{Appendix}
\label{sec:appendix}
\subsection{Training Architectures and Parameters}
\label{subsec:trainParam}
\begin{figure}[h]

\centering
\begin{subfigure}[b]{0.49\columnwidth}
  \centering
	{\includegraphics[width=0.99\columnwidth]{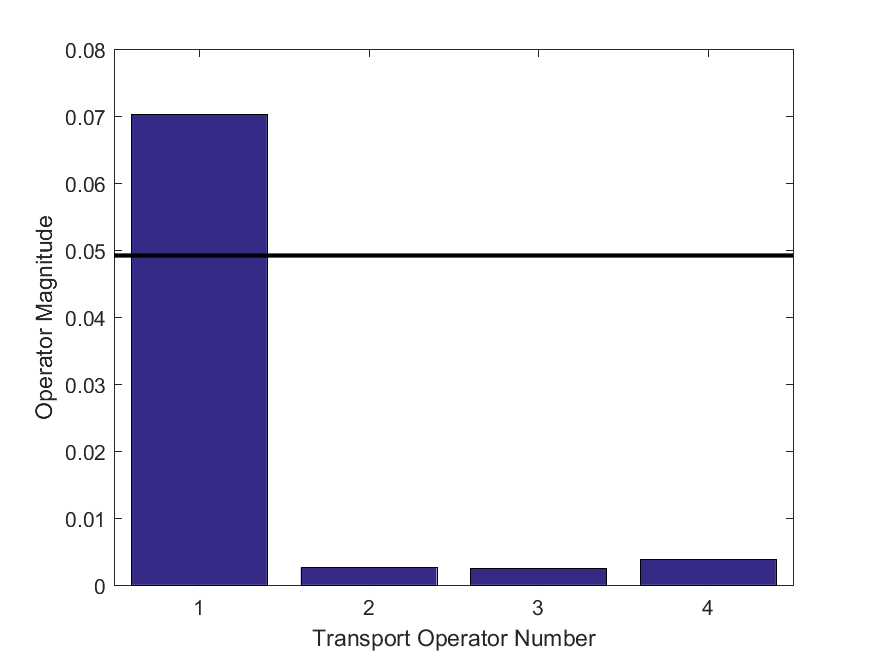}}
  \caption{}
	\label{fig:circleTOMag}
\end{subfigure}
\begin{subfigure}[b]{0.49\columnwidth}
  \centering
	{\includegraphics[width=0.99\columnwidth]{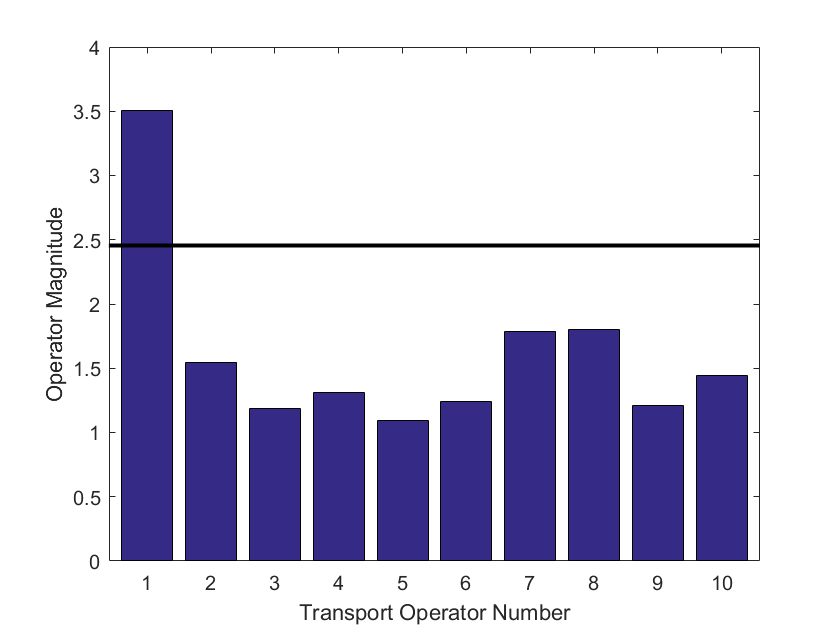}}
  \caption{}
	\label{fig:rotTOMag}
\end{subfigure}
\begin{subfigure}[b]{0.49\columnwidth}
  \centering
	{\includegraphics[width=0.99\columnwidth]{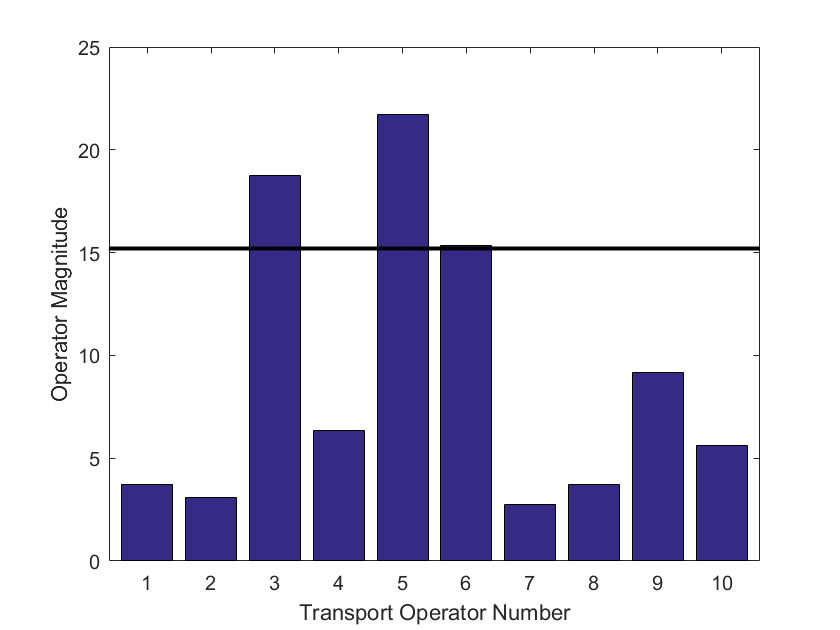}}
	\caption{}
	\label{fig:gaitTOMag}
\end{subfigure}

  \caption{\label{fig:TOMag} Magnitude of the operators after the transport operator training phase. (a) For 2D circle experiment (b) For the rotated MNIST dataset (c) For gait sequences.}
	
\end{figure}

All experiments were run in pytorch. In 2D circle experiment, we used single-layer neural networks with 512 hidden units and ReLU nonlinearities for both the encoder and decoder. Fig.~\ref{fig:circleTOMag} shows the magnitude of transport operators after the transport operator training phase. We select a threshold that is $70\%$ of the magnitude of the transport operator with the maximum magnitude. In this setting the threshold is 0.0492 and only transport operator 1 surpassed the threshold. The training parameters are shown in Table~\ref{tab:circleParams}.

\begin{table}[h]
\centering

\begin{tabular}{||l l l||} 
 \hline
 Autoencoder & Transport Operator  & Fine-tuning  \\ 
 Training & Training & \\
 \hline
 batch size: 64 & batch size: 10 & batch size: 10  \\ 
 steps: 3000 & steps: 3000 & steps: 100  \\
 $lr_{\phi}: 0.0001$ & $lr_{\phi}:$ - & $lr_{\phi}: 0.0001$ \\
 $lr_{\Psi}:$ - & $lr_{\Psi}: 5$  & $lr_{\Psi}: 0.0005$ \\ 
 $\zeta:$ - & $\zeta: 0.0001$  & $\zeta: 0.0$ \\
 $\gamma:$ - & $\gamma: 0.005$  & $\gamma: 0.0$ \\
 $\lambda:$ - & $\lambda:$ - & $\lambda: 1000$ \\
 $M:$ - & $M: 4$ & $M: 1$ \\
 \hline
\end{tabular}
\caption{Training parameters for 2D circle experiment}
\label{tab:circleParams}
\end{table}

For training on MNIST digits we select 50,000 training digits from the traditional MNIST training set and save the additional 10,000 images for validation. We use the traditional MNIST test dataset for testing. We pre-process the MNIST digits by scaling the pixel values between 0 and 1. During transport operator training, we scale the latent vectors prior to performing coefficient inference in order to maintain a maximum value of about 1 for the latent vectors. The network architectures are given in Table~\ref{tab:rotDigitNet} and the training parameters are given in Table~\ref{tab:rotDigitParams}. Fig.~\ref{fig:rotTOMag} shows the magnitude of transport operators after the transport operator training phase with the threshold for transport operator selection. In this experiment the threshold is 2.4556 and only transport operator 1 surpasses this threshold.

\begin{table}[t]
\centering
\begin{tabular}{||l l||} 
 \hline
 Encoder Network & Decoder Network  \\ 
 \hline
 Input $\in \mathbb{R}^{28 \times 28}$ & Input $\in \mathbb{R}^2$  \\ 
 conv: chan: 64 , kernel: 4,  & Linear: 3136 Units  \\
 stride: 2, pad: 1 & \\
 ReLU & ReLU \\
 conv: chan: 64 , kernel: 4,  & convTranpose: chan: 64 ,   \\ 
 stride: 2, pad: 1 & kernel: 4, stride: 1, pad: 1 \\
 ReLU & ReLU \\
 conv: chan: 64 , kernel: 4,  & convTranpose: chann: 64 ,   \\ 
 stride: 2, pad: 0 & kernel: 4, stride: 2, pad: 2 \\
 ReLU & ReLU \\
 Linear: 2 Units & convTranpose: chan: 1 ,   \\ 
  & kernel: 4, stride: 2, pad: 1 \\
   & tanh \\ 
 \hline
\end{tabular}
\caption{Autoencoder network architecture for rotated MNIST experiment}
\label{tab:rotDigitNet}
\end{table}

\begin{table}[t]
\centering
\begin{tabular}{||l l l||} 
 \hline
 Autoencoder & Transport Operator & Fine-tuning  \\ 
 Training & Training & \\
 \hline
 batch size: 64 & batch size: 32 & batch size: 32  \\ 
 epochs: 25 & steps: 2250 & steps: 7800  \\
 $lr_{\phi}: 0.0001$ & $lr_{\phi}:$ - & $lr_{\phi}: 0.005$ \\
 $lr_{\Psi}:$ - & $lr_{\Psi}: 0.01$  & $lr_{\Psi}: 1$ \\ 
 $\zeta:$ - & $\zeta: 0.01$  & $\zeta: 0.0$ \\
 $\gamma:$ - & $\gamma: 8e-5$  & $\gamma: 0.0$ \\
 $\lambda:$ - & $\lambda:$ - & $\lambda: 10$ \\
 $M:$ - & $M: 10$ & $M: 1$ \\
 \hline
\end{tabular}
\caption{Training parameters for rotated MNIST experiment}
\label{tab:rotDigitParams}
\end{table}

We use walking sequences from subject 35 in the CMU Graphics Lab Motion Capture Database in the gait experiment. We use sequences 1-16 for training, sequences 28 and 29 for validation, and sequences 30-34 for testing. During transport operator training, we scale the latent vectors prior to performing coefficient inference in order to maintain a maximum value of about 1 for the latent vectors. Fig.~\ref{fig:gaitTOMag} shows the magnitude of transport operators after the transport operator training phase with the treshold for transport operator selection. The threshold in this experiment is 15.197 and transport operators 3, 5, and 6 all have magnitudes above that threshold. In this case, the transport operators from the transport operator training phase generate full gait sequences so we use these operators for our experiments without undergoing the fine-tuning phase. The network architectures are given in Table~\ref{tab:gaitNet} and the training parameters are given in Table~\ref{tab:gaitParams}. 

\begin{table}[t]
\centering
\begin{tabular}{||l l||} 
 \hline
 Encoder Network & Decoder Network  \\ 
 \hline
 Input $\in \mathbb{R}^{50}$ & Input $\in \mathbb{R}^5$  \\ 
 Linear: 512 Units & Linear: 512 Units  \\
 tanh & tanh \\
 Linear: 512 Units & Linear: 512 Units  \\
 tanh & tanh \\
 Linear: 512 Units & Linear: 512 Units  \\
 tanh & tanh \\
 Linear: 5 Units & Linear: 50 Units \\ 
 \hline
\end{tabular}
\caption{Autoencoder network architecture for walking gait experiment}
\label{tab:gaitNet}
\end{table}

\begin{table}[t]
\centering
\begin{tabular}{||l l ||} 
 \hline
 Autoencoder Training & Transport Operator Training  \\ 
 \hline
 batch size: 64 & batch size: 32   \\ 
 training steps: 15000 & training steps: 14500   \\
 $lr_{\phi}: 0.0005$ & $lr_{\phi}:$ -  \\
 $lr_{\Psi}:$ - & $lr_{\Psi}: 0.005$   \\ 
 $\zeta:$ - & $\zeta: 0.05$   \\
 $\gamma:$ - & $\gamma: 0.0001$   \\
 $\lambda:$ - & $\lambda:$ -  \\
 $M:$ - & $M: 10$  \\
 \hline
\end{tabular}
\caption{Training parameters for walking gait experiment}
\label{tab:gaitParams}
\end{table}

\begin{figure}[ht]

\centering
\begin{subfigure}[b]{0.69\columnwidth}
  \centering
	{\includegraphics[width=0.95\columnwidth]{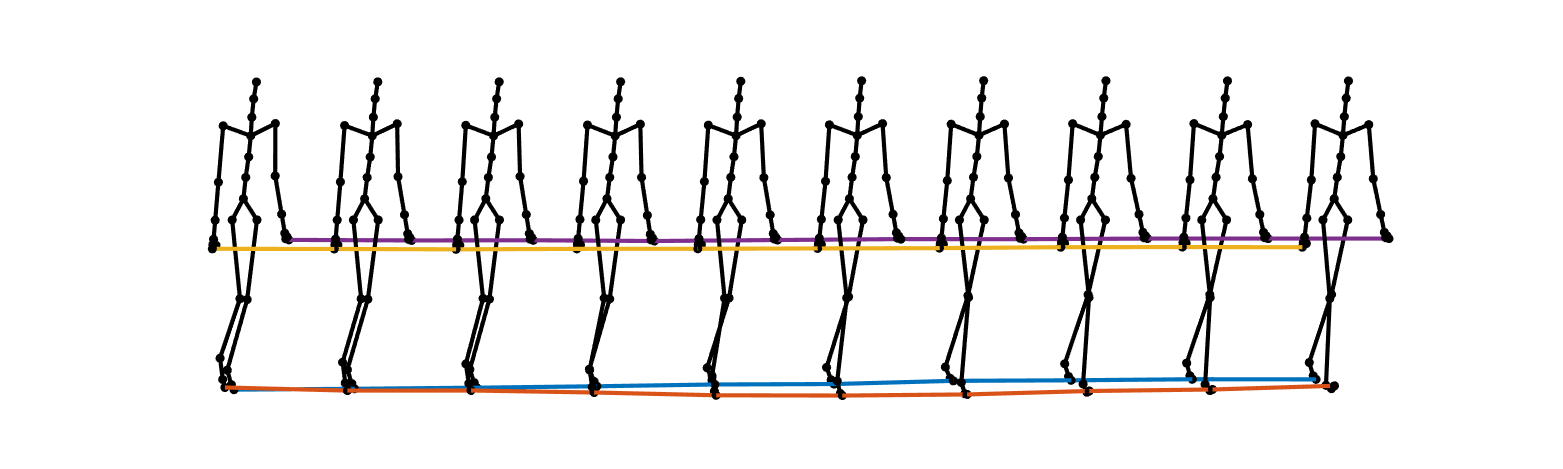}}
  \caption{}
	\label{subfig:GaitSeq3}
\end{subfigure}
\begin{subfigure}[b]{0.69\columnwidth}
  \centering
	\includegraphics[width=0.95\columnwidth]{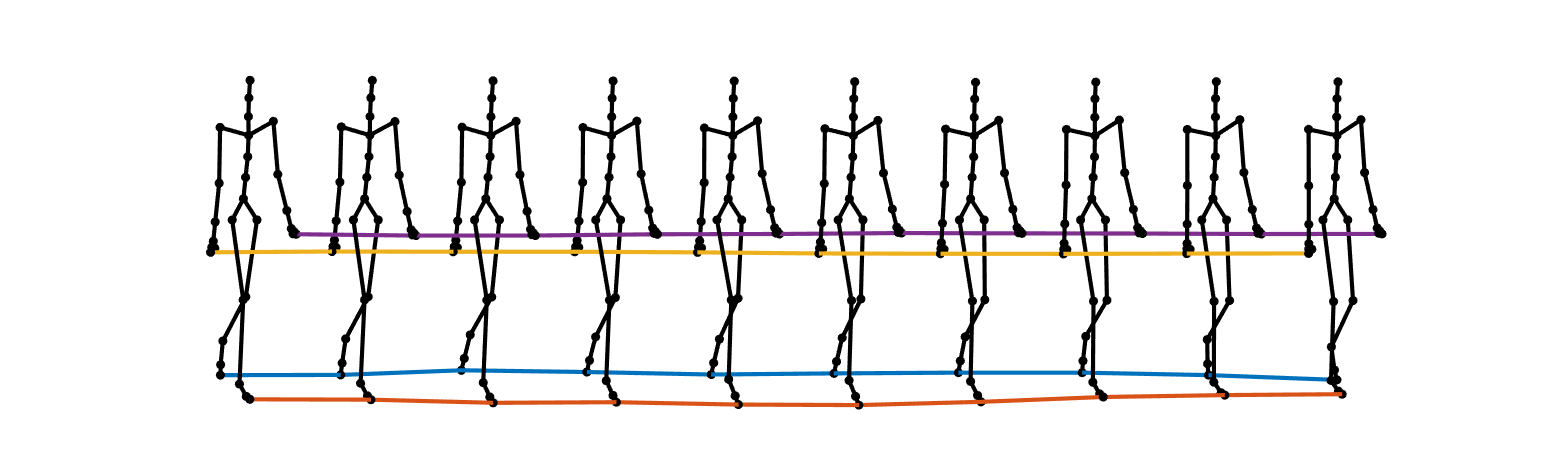}
	\caption{}
	\label{subfig:GaitSeq6}
\end{subfigure}

  \caption{\label{fig:gaitSeq} (a) Gait sequence generated from operator 3. (b)  Gait sequence generated from operator 6. }
	
\end{figure}

\begin{figure}[ht]
\centering
\begin{subfigure}[b]{0.45\columnwidth}
  \centering
	{\includegraphics[width=0.95\columnwidth]{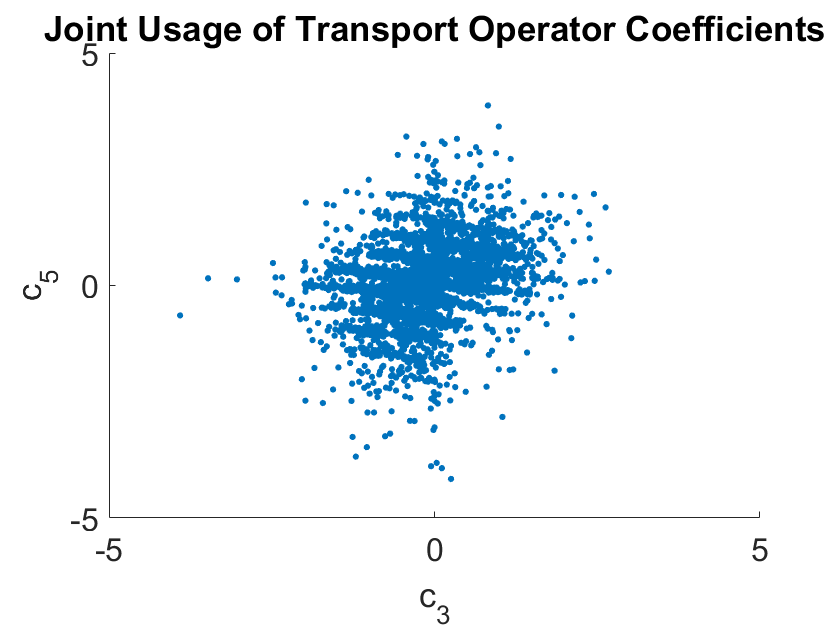}}
  \caption{}
	\label{subfig:jointCoeff35}
\end{subfigure}
\begin{subfigure}[b]{0.45\columnwidth}
  \centering
	{\includegraphics[width=0.95\columnwidth]{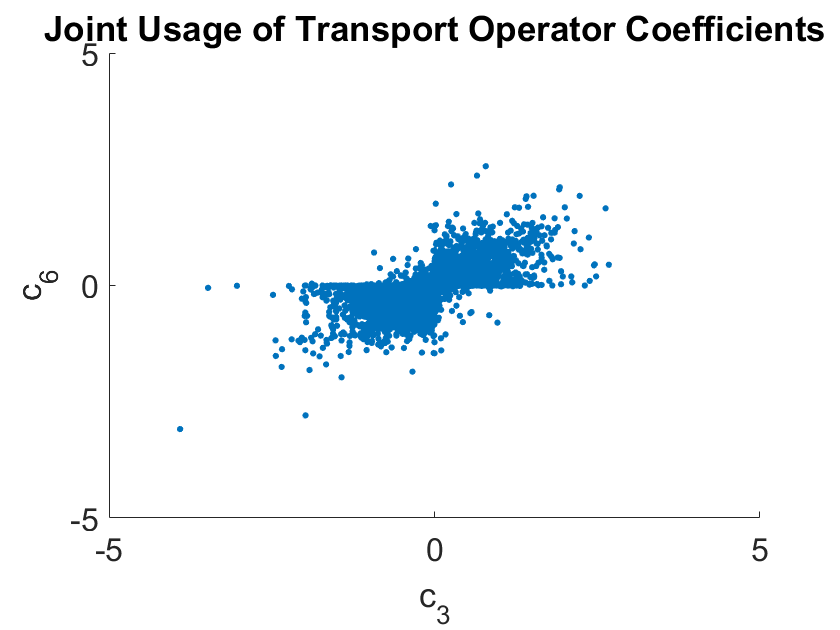}}
  \caption{}
	\label{subfig:jointCoeff36}
\end{subfigure}
\begin{subfigure}[b]{0.45\columnwidth}
  \centering
	\includegraphics[width=0.95\columnwidth]{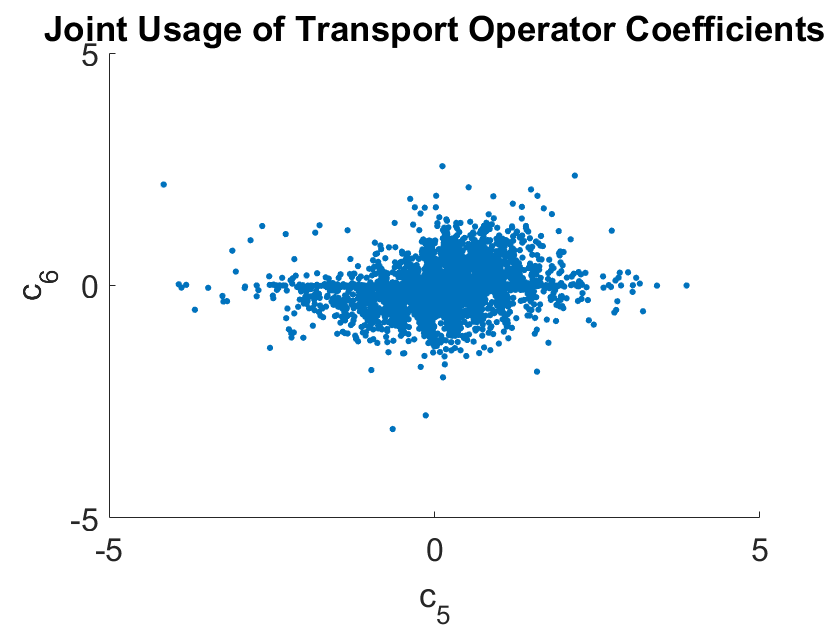}
	\caption{}
	\label{subfig:jointCoeff56}
\end{subfigure}

  \caption{\label{fig:jointCoeff} Scatter plots of the inferred coefficients for pairs of transport operators. The coefficients were inferred over the latent representations of pairs of points in the gait sequences. (a) Coefficients for transport operator 3 and transport operator 5. (b) Coefficients for transport operator 3 and transport operator 6. (c) Coefficients for transport operator 5 and transport operator 6.}
	
\end{figure}

\subsection{Analysis of Joint Usage of Gait Transport Operators}\label{sec:joinGait}
The gait experiment is the only experiment that results in more than one dictionary element above the selection threshold. To better understand the usage of the operators, this section shows a visualization of the action of each of these operators and how their coefficients are activated jointly during gait sequences. Fig~\ref{fig:walkTransOpt} shows the gait sequence generated by transport operator 5. Fig.~\ref{fig:gaitSeq} shows the gait sequences generated from the two remaining operators above the selection threshold. All three operators induce continuous gait sequences but they vary in speed and mechanics of the gait. For instance, transport operator 5, shown in Fig.~\ref{fig:walkTransOpt}, generates a faster gait sequence than the other two and the body rocks side to side more with that operator. Transport operator 3 (shown in Fig.~\ref{subfig:GaitSeq3}) on the other hand results smaller steps than the other two with the body tilted further forward.

Fig.~\ref{fig:jointCoeff} shows plots of the inferred coefficients for pairs of each of the three high magnitude transport operators. These plots show that the transport operators are used jointly and there are shared patterns of usage between them.

\subsection{Additional Examples of Transport Operators}\label{sec:exTO}
\begin{figure}[ht]

\centering
\begin{subfigure}[b]{0.62\columnwidth}
  \centering
	{\includegraphics[width=0.99\columnwidth]{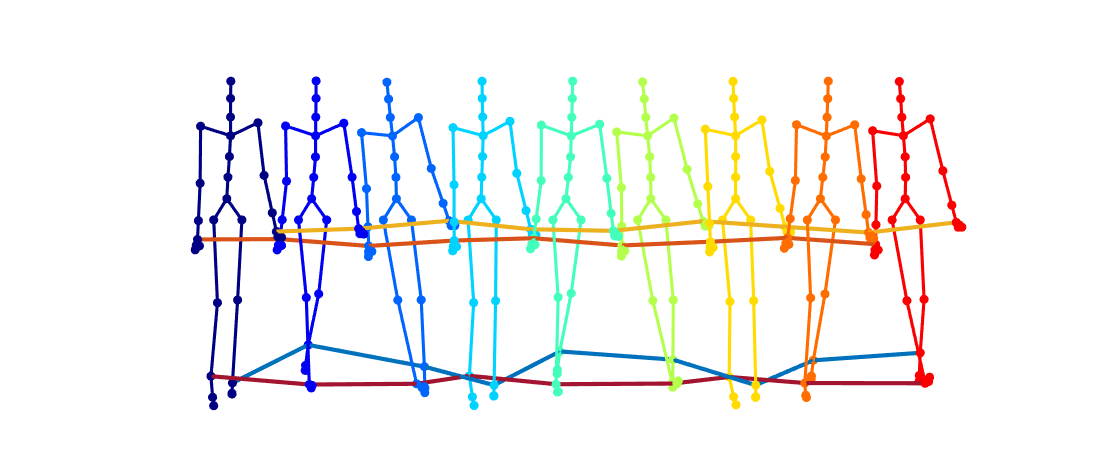}}
  \caption{}
	\label{subfig:inputGait1}
\end{subfigure}
\begin{subfigure}[b]{0.62\columnwidth}
  \centering
	{\includegraphics[width=0.99\columnwidth]{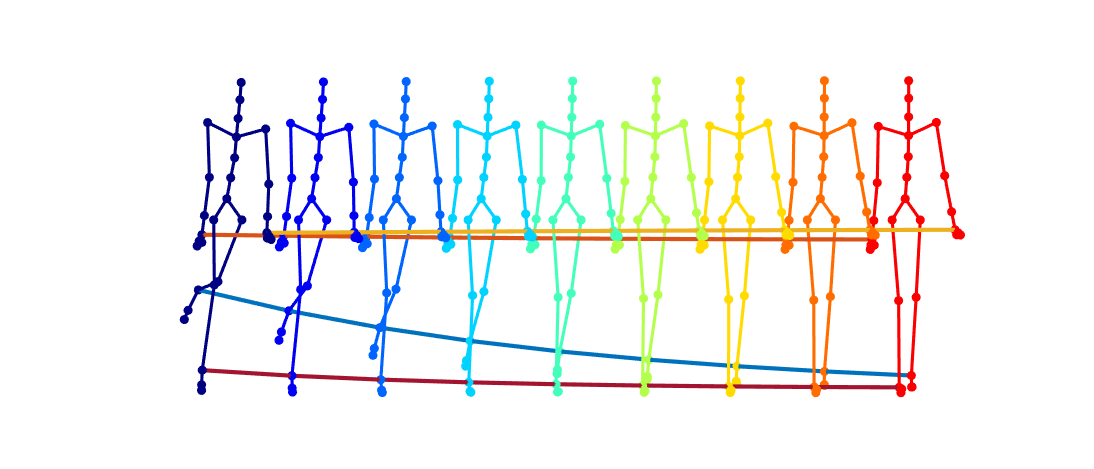}}
  \caption{}
	\label{subfig:inputGait2}
\end{subfigure}
\begin{subfigure}[b]{0.62\columnwidth}
  \centering
	\includegraphics[width=0.99\columnwidth]{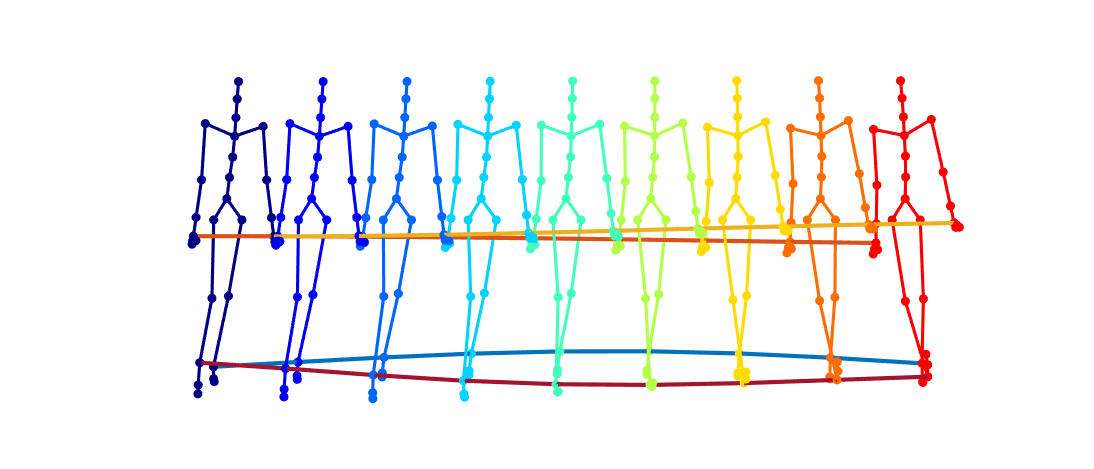}
	\caption{}
	\label{subfig:inputGait4}
\end{subfigure}
\begin{subfigure}[b]{0.62\columnwidth}
  \centering
	\includegraphics[width=0.99\columnwidth]{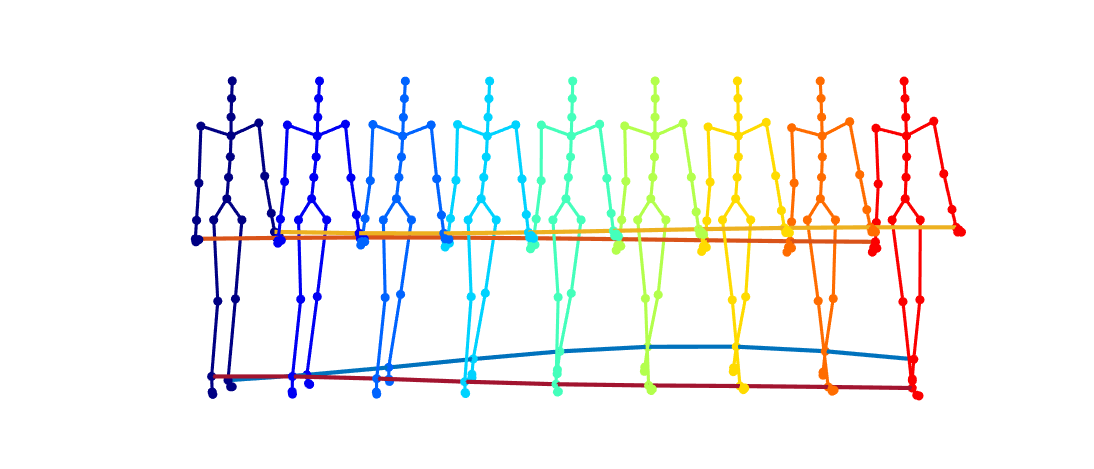}
	\caption{}
	\label{subfig:inputGait8}
\end{subfigure}

  \caption{\label{fig:inputGait} The effect of applying four example transport operators trained on gait data in the input space. }
	
\end{figure}

To provide more context to the transport operators shown in this paper, we show transport operators trained with the CMU Graphics Lab Motion Capture data in the input space, rather than in the latent space of an autoencoder. Multiple operators are learned that encompass different movements. Fig.~\ref{subfig:inputGait1} shows the effect of an operator that induces a walking sequence. Fig.~\ref{subfig:inputGait2} shows the effect of an operator that kicks the left foot backward. Fig.~\ref{subfig:inputGait4} shows the effect of an operator that causes the body to lean from side-to-side during the gait sequence.

\subsection{Hyperspherical VAE}\label{sec:hvae}
Our hyperspherical VAE implementation came from the Nicola De Cao's github page.\footnote{\url{https://github.com/nicola-decao/s-vae-pytorch/tree/master/hyperspherical_vae}} For both the concentric circle dataset and the gait sequences dataset, we trained the hyperspherical VAE with the same network architectures as our autoencoder experiments and a mean square error reconstruction loss. For the rotated MNIST dataset, we used the network architecture from the mnist example in the hyperspherical VAE code and used the binary cross entropy loss for the reconstruction error on dynamically binarized rotated digit images.

To estimate paths on the hyperspherical VAE latent space, we used the Manifold-valued Image Restoration Toolbox\footnote{\url{https://ronnybergmann.net/mvirt/}} to compute geodesic paths on a 10-dimensional hypersphere. 

\small
\bibliography{AAAI-ConnorM.4191.bib}
\bibliographystyle{aaai}

\end{document}